\documentclass{article}

\usepackage{microtype}
\usepackage{graphicx}
\usepackage{subfigure}
\usepackage{booktabs} 

\usepackage{hyperref}

\usepackage[accepted]{icml2020}
\usepackage{amsmath}
\usepackage{amssymb}
\usepackage{longtable}

\usepackage{enumitem}
\setlist{nosep} 

\usepackage{multirow}
\usepackage{enumitem}
\usepackage{pifont}
\usepackage[dvipsnames]{xcolor}

\newcommand{\minisection}[1]{\vspace{-2pt} \noindent {\bf #1}\ \ }

\icmltitlerunning{Bookworm continual learning: beyond zero-shot learning and continual learning}

\begin{document}
	
	\twocolumn[
	
	\icmltitle{Bookworm continual learning:\\ beyond zero-shot learning and continual learning}
	
	
	
	\icmlsetsymbol{equal}{*}
	
	\begin{icmlauthorlist}
		\icmlauthor{Kai Wang}{cvc}
		\icmlauthor{Luis Herranz}{cvc}
		\icmlauthor{Anjan Dutta}{exeter}
		\icmlauthor{Joost van de Weijer}{cvc}
	
	\end{icmlauthorlist}
	
	\icmlaffiliation{cvc}{Computer Vision Center, Autonomous University of Barcelona (UAB)}
	\icmlaffiliation{exeter}{Computer Vision and Machine Learning, University of Exeter}

	\icmlcorrespondingauthor{Luis Herranz}{lherranz@cvc.uab.es}
	
	
	\icmlkeywords{Machine Learning, ICML}
	
	\vskip 0.3in
	
	]
	
	
\printAffiliationsAndNotice{}  

\begin{abstract}
We propose \textit{bookworm continual learning} (BCL), a flexible setting where unseen classes can be inferred via a semantic model, and the visual model can be updated continually. Thus BCL generalizes both continual learning (CL) and zero-shot learning (ZSL). We also propose the \textit{bidirectional imagination} (\textit{BImag}) framework to address BCL where features of both past and future classes are generated. We observe that conditioning the feature generator on attributes can actually harm the continual learning ability, and propose two variants (joint class-attribute conditioning and asymmetric generation) to alleviate this problem. 
\end{abstract}


\section{Introduction}

Deep learning has brought extraordinary success to visual recognition by learning from large amounts of data (e.g. object classification and detection, scene classification). There are, however, two critical assumptions that stem from a \textit{static} view of the world: all concepts of interest are known before training, and the corresponding training data is also available beforehand. The resulting model is also static and remains unchanged after training. Another limitation of conventional classification models is that there is no explicit notion of semantic similarity between concepts (i.e. a \textit{semantic model}), since classes are represented as one-hot labels (i.e. all classes are equally similar and dissimilar to each other). These assumptions are hardly met in the \textit{dynamic} real world we live in, where new visual data and new semantic concepts are continuously observed and integrated in our own personal knowledge. Similarly, visual recognition in humans greatly leverages all sort of semantic (and contextual) knowledge, enabling sophisticated inference.

\begin{figure}[t]
	\setlength{\tabcolsep}{3pt}
		\centering
		
			\vskip 0.2in
				\centerline{\includegraphics[width=\columnwidth]{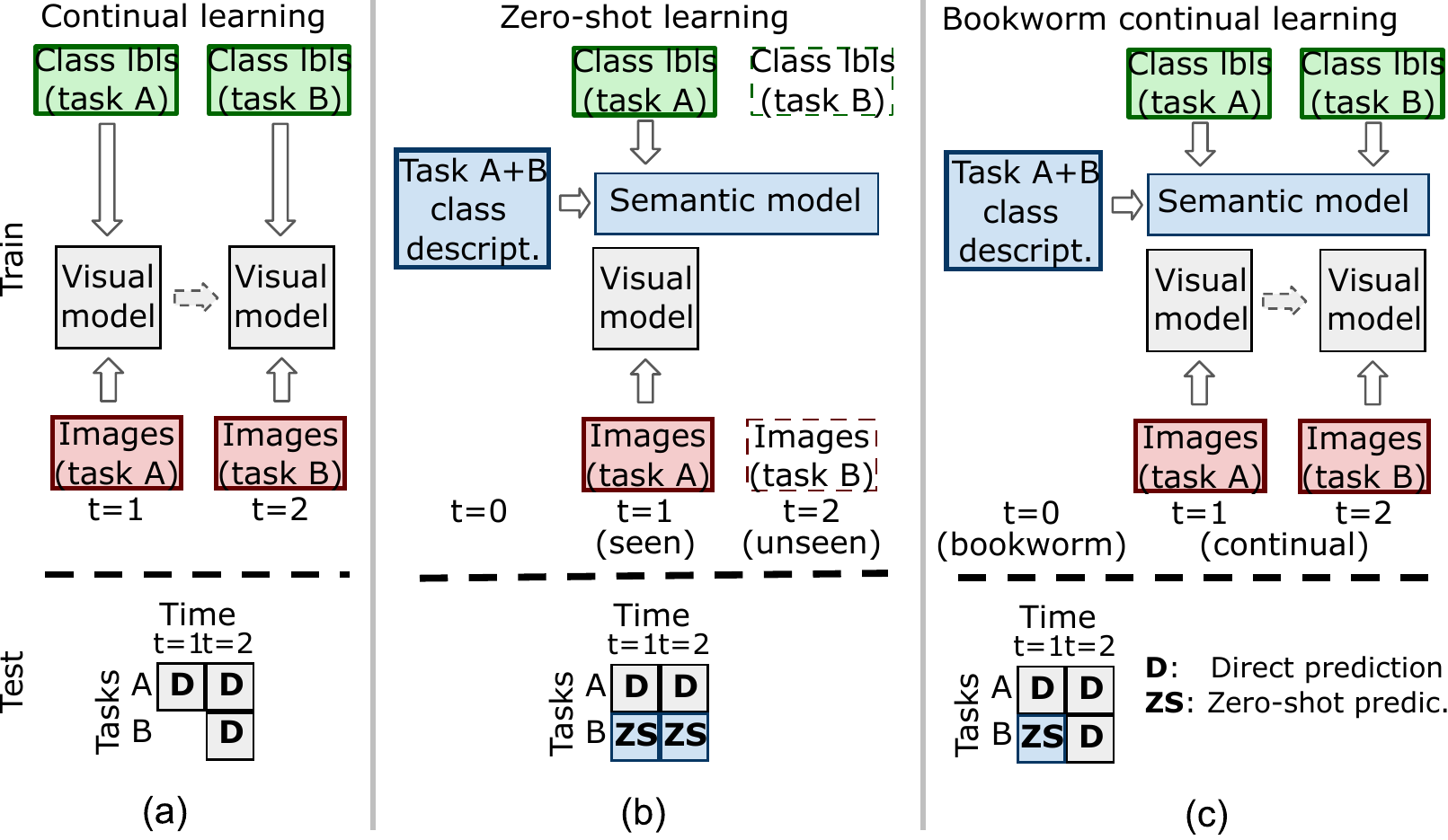}}
				\label{fig:generalized-continual-learning}
				\caption{Generalized continual learning: (a) continual learning, (b) zero-shot learning, and (c) bookworm continual learning.}
			\vskip -0.3in
	\end{figure}

Challenging this static world assumption, continual learning (CL) focuses on how to update the visual model when new classes and visual instances are observed over time (see Fig.~\ref{fig:generalized-continual-learning}a). A consequence is that the data is no longer i.i.d. and learning new tasks results in forgetting previous ones (i.e. catastrophic forgetting). This problem has been addressed with different techniques, including weight regularization~\cite{Kirkpatrick2017EWC, Aljundi2018MAS, Liu2018RotateNet} distillation~\cite{Li2017LwF}, episodic memories with exemplars~\cite{Rebuffi2017iCarl} and generative replay methods~\cite{Shin2017DGR,Chenshen2018MeRGAN}.
	    
On the other hand, zero-shot learning (ZSL) enables the recognition of (visually) unseen classes via a semantic model that describes them in connection to the seen classes (see Fig.~\ref{fig:generalized-continual-learning}b). We can also observe that ZSL also has an implicit temporal structure, with the class descriptions learned first, then the visual model is learned from the data of seen classes, and then the model is tested over the unseen classes. ZLS is usually tackled as learning the alignment between visual features and class embeddings (via the semantic model) in an shared intermediate space~\cite{frome2013devise,akata2015label}. Recent works also use feature generators to synthesize features of unseen classes~\cite{xianCVPR18,Mishra_2018_CVPR_Workshops,XianCVPR2019a}.

In this work, we argue that continual leaning and semantic models are both essential for advanced visual recognition. Therefore, we propose \textit{generalized continual learning} (GCL) as a more realistic setting where visual recognition is addressed with the help of an explicit semantic model, and in a dynamic scenario that requires continual learning. In the rest of the paper we focus on a particular case that we refer to as \textit{bookworm\footnote{We use an avid reader (i.e.~\textit{bookworm} stereotype) as a metaphor, due to his/her extensive encyclopedic knowledge (e.g. concept descriptions) before eventually observing them visually.} continual learning} (BCL) where the semantic model remains fixed while the visual model is updated continuously (see Fig.~\ref{fig:generalized-continual-learning}c). BCL can be seen as a generalization of CL which is limited by lacking explicit semantic models, and ZSL which is not continual. The main challenge of GCL is the effective integration of semantic models and CL.

We propose a unified BCL framework via feature generation and distillation. 
A generative model (a conditional VAE) learns the distribution of features of past and future classes and generates synthetic features so a joint classifier on all classes can be trained. In our first BCL model, the feature generator is conditioned on attributes (\textit{attr-BImag} variant).

We further observe that conditioning on attributes severely hurts the ability of the feature generator to prevent forgetting, compared to its continual learning counterpart. This raises the question of whether attributes are helpful or harmful in dealing with forgetting. We further investigate the problem, noticing an asymmetry between backward and forward generation (past classes have been visually observed, but not future ones), and inherent limitations of attribute-based semantic models themselves. Addressing these limitations we propose three variants with improved performance, while being also memory and computationally efficient. Finally, we also propose a novel metric to evaluate BCL, which generalizes a GZSL metric, not used earlier to evaluate CL.

\section{Bookworm continual learning}
\subsection{Bookworm and generalized continual learning}
We assume a sequence of image classification tasks $\left(S_1,\ldots,S_K\right)$. Each task is learned from a dataset $\mathcal{S}_k=\left\{\left(\mathbf{x}_i^k,\mathbf{a}_i^k,y_i^k\right)_{i=1}^{N_k}\right\}$, where $\mathbf{x}_i^k\in \mathcal{X}_k$ is an image, $y_i^k\in \mathcal{Y}_k\subset \mathcal{Y}$ is the corresponding class label and $\mathbf{a}_i^k\in \mathcal{A}_k\subseteq \mathcal{A}$ is the semantic description. We are ultimately interested in learning and continually updating a visual model $p_t\left(y\vert \mathbf{x}\right)=C_t\left(F_t\left(\mathbf{x}\right)\right)$ that maps images to class probabilities, where $\mathbf{z}=F_t\left(\mathbf{x}\right)$ and $p_t\left(y\vert \mathbf{z}\right)=C_t\left(\mathbf{z}\right)=\textnormal{softmax}\!\left(W_t^\intercal\mathbf{z}\right)$ are the visual feature extractor and the classifier at time $t$, respectively (all implemented jointly as a deep neural network). For simplicity, we assume that $k$-th task is learned at time $t=k$ and will use $t$ and $k$ interchangeably.

We also consider a semantic model $p\left(y\vert \mathbf{a}\right)$ ) that relates class and attributes\footnote{For simplicity, we assume classification tasks and attribute-based semantic models, but our discussion is also valid for any other fixed-size continuous semantic embeddings (e.g. word embeddings, language embeddings).}. The semantic model is learned or annotated from an external source (e.g. class descriptions, taxonomy, Wikipedia), and can be leveraged to help infer classes, including unseen ones, whose instances might have not been observed yet (but their descriptions have). The visual model is always updated over time. In GCL the semantic model can be also continually updated, while in BCL it is learned prior to the visual model during a \textit{bookworm} stage (at $t=0$, for simplicity). We focus on the latter in this paper (see Fig.~\ref{fig:generalized-continual-learning}c), and assume \textit{task-agnostic} evaluation, i.e. during test the task is unknown and the model has to consider all classes for the prediction.

Zero-shot learning (ZSL) can be seen as the particular case of BCL with two tasks and no update after the first one. Using ZSL terms, the first task is \emph{seen} and the second is \emph{unseen}, i.e. $\mathcal{Y}_1=\mathcal{Y}_\text{seen}$, $\mathcal{Y}_2=\mathcal{Y}_\text{unseen}$. The model is evaluated on $\mathcal{Y}_\text{seen}$, which can be inferred using the semantic model. Generalized ZSL (GZSL) corresponds to task-agnostic evaluation, i.e. over $\mathcal{Y}_\text{seen}\bigcup \mathcal{Y}_\text{unseen}$. Continual learning (CL) corresponds to the particular case where no semantic model is available, and therefore at time $t$ the model can only discriminate between all the classes seen so far, which we denote as $\mathcal{Y}_{\leq t}=\bigcup_{k=1}^t \mathcal{Y}_k$. Finally, if we further assume no continual update we recover the usual setting where the model is learned with all the data $\mathcal{S}=\bigcup_k \mathcal{S}_k$ (we refer to it as \textit{joint training} (JT)).

\section{BImag: feature generation for BCL}

\begin{figure}[t]
	\setlength{\tabcolsep}{3pt}
	\centering
	\vskip 0.2in
	\begin{center}
		\centerline{\includegraphics[width=\columnwidth]{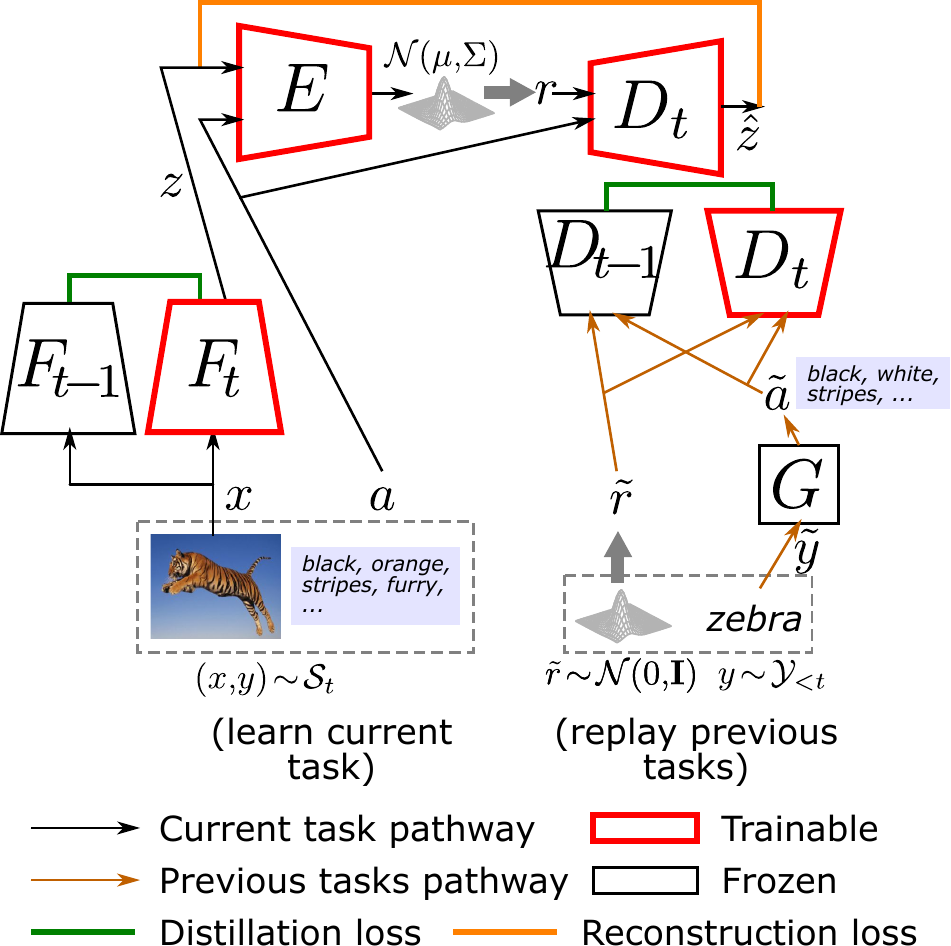}}		
		\caption{Bidirectional imagination framework (\textit{attr-BImag}).\label{fig:bimag}}
	\end{center}
	\vskip -0.4in
\end{figure}

\subsection{Integrating continual learning and semantic models}
To address BCL we need to cope with three challenges: (a) \textit{catastrophic interference} between tasks in the shared feature extractor, (b) \textit{bias} in the classifier (to the most recent observed data), and (c) a way to \textit{predict future classes} (via semantic information). Here (a) is related to CL, (c) to GZSL and (b) to both.
Our approach tackles these challenges separately with distillation in the feature extractor to prevent catastrophic interference, and synthetic data generation to train a joint and unified classifier for all classes (see Fig.~\ref{fig:bimag}). Compared to traditional generative replay in CL~\cite{Shin2017DGR}, our method focuses only on the classifier generating features rather than images, leverages semantic information, has a hierarchical generator which is also bidirectional (generates features of past, i.e. \textit{replay}, and future classes, i.e. \textit{foresight/imagination}), and hence we loosely refer to our framework as \textit{bidirectional imagination} (\textit{BImag}, see Fig.~\ref{fig:bimag}). This allows us to predict any category at any time, while also allowing for continual updates. Semantic information is only used during training, and the final model is a direct mapping from image to class, without mapping to any intermediate semantic space. 

\minisection{Overview}
In a first stage to learn a new task at time $t$, the feature extractor $F_t$ is updated using an auxiliary classifier $\hat{C}_t$ minimizing the cross-entropy loss over the current task. Forgetting is alleviated by distilling the features of a fixed copy of the previous feature extractor $F_{t-1}$ to the current feature extractor $F_t$, using $l_2$ loss and computed over the images of the current task.

In the second stage (see Fig.~\ref{fig:bimag}a) we train a conditional variational autoencoder  with an encoder $\left[\mathbf{\mu},\Sigma\right]=E\left(\mathbf{z},\mathbf{a}\right)$ (which estimates the parameters of the multivariate Gaussian latent distribution), and a decoder $D_t\left(\mathbf{r},\mathbf{a}\right)$, and $\mathbf{r}$ is a random latent vector (i.e. $\mathbf{r}\sim \mathcal{N}\left(\mathbf{\mu},\Sigma\right)$). The description generator $\mathbf{a}=G\left(y\right)=A\mathbf{1}_y$ maps the class $\mathbf{y}$ to the attribute-based description $\mathbf{a}$ via the class-to-attribute matrix $A$\footnote{$\mathbf{1}_y$ represents the one-hot representation of $y$}. The decoder will act as feature generator conditioned on the attribute-based description. 
The parameters of the encoder and decoder are learned by maximizing the evidence lower bound (ELBO). The feature extractor remains fixed during this stage, and the encoder is learned from scratch every time. In addition, we include the replay alignment loss~\cite{Chenshen2018MeRGAN} between the past decoder $D_{t-1}$ and the current decoder $D_t$, which is a form of distillation to prevent forgetting in the feature generator.

Once the VAE is trained, the decoder can generate a set of synthetic features $\tilde{\mathcal{S}}_{\neq t}$ for both past and future classes. The classifier $C_t$ is trained with both real and synthetic features, i.e. $\mathcal{S}_t\bigcup \tilde{\mathcal{S}}_{\neq t}$ using the cross-entropy loss. We refer to this variant with attribute-conditional VAE as \textit{attr-BImag}.

\minisection{Relation to CL}
We notice that the VAE of \textit{attr-BImag} with $A=\mathbb{I}$ is directly conditioned on the class label, resulting in a CL framework because it cannot predict future classes, which we use as CL baseline (i.e. \textit{class-BImag}).

Interestingly, we observed that in practice \textit{attr-BImag} tends to forget more than \textit{class-BImag} (i.e. attributes, rather than helping, are harming the ability to prevent forgetting previous ones). In order to understand this problem, it is convenient to observe that feature generation in \textit{class-BImag} can be formulated as $\mathbf{z}\sim p\left(\mathbf{z} \vert y\right)$. Similarly, we can add attributes as another variable and factorize as $p\left(\mathbf{z} \vert y\right)=p\left(\mathbf{z} \vert \mathbf{a}, y\right)p\left(\mathbf{a} \vert y\right) \label{eq:factorized_model}$.

The particular case of \textit{attr-BImag} computes $\mathbf{a}=G\left(y\right)$, followed by sampling $\mathbf{z}\sim p\left(\mathbf{z}\vert\mathbf{a}\right)$. Thus, \textit{attr-BImag} assumes that features and classes are independent, and therefore all relevant visual information to generate synthetic features needs to be represented somehow in the attribute space. This is difficult to achieve in practice, and the feature generation may be unable to synthesize certain discriminative patterns that are essential to keep high accuracy and prevent forgetting. In contrast, \textit{class-BImag} has a direct mapping between classes and features, so the feature generator could, in principle, model directly the relevant visual information and capture its diversity.

\minisection{Joint class-attribute conditioning}
We can partly alleviate the dependence on the attribute space by conditioning the VAE both on attributes and classes, and then generate features as $\mathbf{z}\sim p\left(\mathbf{z}\vert\mathbf{a}, y\right)$ (see \ref{eq:factorized_model}). We refer to this variant as \textit{class-attr-BImag}.

\minisection{Asymmetric generation}
Feature generation in BImag is asymmetric: at a given time, the feature generator has observed only the semantic description of future classes, while has observed both visual and semantic information of past classes. As we discussed previously, conditioning directly on visual information seems to prevent forgetting better than conditioning on attributes, but the latter is necessary to predict unseen classes. Motivated by this observation we decouple both generation directions and use a different VAE for each (\textit{asym-BImag}), one conditioned on classes for backward generation and the other conditioned on attributes for forward generation.

\section{Experiments}

\subsection{Settings}

\minisection{Datasets and splits.} CUB is a fine-grained recognition dataset with 200 classes~\cite{wah2011caltech}, while AwA (specifically AwA2) has 50 coarser classes~\cite{xian2018zero}. We follow the settings and preprocessing used in conventional GZSL methods. We use the data, class splits and train/test splits proposed by~\cite{xian2018zero}, adapting them to our BCL setting. This results in two tasks A/B\footnote{We use $t=1,2,\ldots$ to index time and $k=A,B,\ldots$ to index tasks. We assume that the $k$-th task is learned at time $t=k$.} with class splits 150/50 for CUB and 40/10 for AwA (tasks A/B in BCL or seen/unseen in ZSL, respectively). Since task B is not trained in ZSL, we created our own train/test splits.

\minisection{Implementation details.} Our implementation is based on PyTorch and trained using NVIDIA GTX 1080Ti GPUs. The feature extractor in our model is a ResNet-101~\cite{he2016deep}, as commonly used in previous works in ZSL, and then fine tuned on every new task as typically done in CL. Our conditional VAE consists of an encoder with three fully connected layers and a decoder with two fully connected layers (see supplementary material for details). The conditions can be attribute vectors and/or class labels as one-hot vectors. To train the joint classifier (Fig.~\ref{fig:bimag}c), we generate 300 synthetic features per class for both past and future classes. We set $\lambda_1=1$, $\lambda_2=0.1$. We use Adam optimizer~\cite{kingma2014adam} with learning rates 0.0001 for the feature extractor and 0.001 both for classifier and VAE.

\minisection{Baselines and variants.} We use \textit{class-BImag} with fine tuned feature extractor, distillation and replay alignment as main CL baseline. We extend this baseline with different semantic models to the BCL variants \textit{attr-BImag},  \textit{class-attr-BImag} and \textit{asym-BImag}. Note that BCL methods at step $t=1$ correspond to GZSL.

\minisection{Metrics.} We adapt the AUSUC metric used in GZSL~\cite{Changpinyo2016ZSL} and use the area under the (per-class) task-accuracy curve (AUTAC) as metric to evaluate BCL. AUSUC was proposed as a more robust metric than the more common harmonic mean of seen and unseen accuracies~\cite{xian2018zero}, which is very sensitive to score calibration. Finally, to evaluate how a particular approach is able to make predictions for any task or class at any time, which is the main objective in BCL, we compute the average AUTAC across time.

\begin{table}[t]
	\centering
	\def\arraystretch{1.0}
	\setlength\tabcolsep{2pt}
	\resizebox{\columnwidth}{!}{%
		\begin{tabular}{ccc|cccc|cccc}
			\hline
			\multirow{2}{*}{Method} & \multirow{2}{*}{Generator} & \multirow{2}{*}{FE} & \multicolumn{4}{c|}{CUB} & \multicolumn{4}{c}{AWA}\\
			& & & A&B & H & AUTAC & A&B & H & AUTAC \\
			\hline
			\multirow{2}{*}{attr-BImag} & VAE & fix & 60.84 & 39.70 & 48.05 & 0.347 & 72.28& \textbf{62.02} & \textbf{66.76} & 0.540 \\
			& VAE & ft & \textbf{77.74} & 41.30 & 53.94 &  0.484 & 73.83 & 59.97 & 66.18  & 0.555\\
			\multirow{2}{*}{cls-attr-BImag} & VAE & fix & 59.28 & 40.97 & 48.45 & 0.349 & 74.20 & 54.18 & 62.63  & 0.453\\
			& VAE & ft & 73.57 & 45.09 & 55.91 & \textbf{0.515} & \textbf{76.93} & 51.40 & 61.63 & \textbf{0.578}\\
			\hline
			Mishra \textit{et al.} & VAE & fix & - & - & 34.5 & - & - & - & 51.2 & - \\
			f-CLSWGAN & GAN & fix& 57.7 & 43.7 & 49.7 &  -& 61.4 & 57.9  & 59.6 & - \\
			\multirow{2}{*}{f-VAEGAN-D2} & VAE, GAN & fix & 60.1 & 48.4 & 53.6 & - & 70.6 & 57.6 & 63.5 & -\\
			& VAE, GAN & ft & 75.6 & \textbf{63.2} & \textbf{68.9} & - & 76.1 & 57.1 & 65.2 & -\\
			\hline
		\end{tabular}
	}
	\vspace{-1pt}
	\caption{\footnotesize Experiments on GZSL (accuracies in \%) and related works using feature generation. In GZSL, A and B refer to seen and unseen classes respectively.}\label{tab:exp_gzsl}
	\vskip -0.1in
\end{table}

\begin{table}[t]
	\centering
	\def\arraystretch{1.0}
	\setlength\tabcolsep{2pt}
	\resizebox{\columnwidth}{!}{%
		\begin{tabular}{c|c|ccc|c|ccc}
			\hline
			& \multicolumn{4}{c|}{CUB 150/50} & \multicolumn{4}{c}{AWA 40/10}\\
			& CL & \multicolumn{3}{c|}{GZSL/BCL} & CL & \multicolumn{3}{c}{GZSL/BCL}\\
			& class & attr & cls-att & asym & class & attr & cls-att & asym\\
			\hline
			$t=1$ (GZSL) & 0.018 & 0.484 & \textbf{0.515} & 0.484 &0.039 & 0.555 & \textbf{0.578} & 0.555\\
			$t=2$ & 0.691 & 0.670 & 0.685 & 0.691 & 0.917 & 0.914 & \textbf{0.923} & 0.917 \\
			Mean & 0.355 & 0.577 & \textbf{0.600} & 0.588 & 0.478 & 0.735 & \textbf{0.750} & 0.736\\
			\hline
		\end{tabular}
	}
	\vspace{-1pt}
	\caption{Two tasks experiments (AUTAC metric) on CUB 150/50, AwA 40/10 and SUN 645/72.}\label{tab:exp_2tasks}
\end{table}

\subsection{Generalized zero-shot learning}
We first evaluate our framework in the GZSL setting (equivalent to BCL at $t=1$). Table~\ref{tab:exp_gzsl} shows the results
\footnote{We average results over 5 runs. Other GZSL methods in the table do not report average results (possibly reporting the best run).} for CUB 150/50 and AwA 40/10, including recent works using feature generators~\cite{Mishra_2018_CVPR_Workshops}, f-CLSWGAN~\cite{xianCVPR18} and f-VAEGAN-D2~\cite{XianCVPR2019a}), with either fixed (\textit{fix}) or fine tuned (\textit{ft}) feature extractor. Although it was not our main objective, BImag achieves very competitive results, including the best result in AwA, and second best in CUB, only behind f-VAEGAN-D2 (ft). Interestingly, conditioning on class label seems to be also beneficial to GZSL. 

\subsection{Bookworm continual learning}
Table~\ref{tab:exp_2tasks} shows the results for two tasks for the different variants of BImag. The CL variant \textit{class-BImag} cannot predict future classes, in contrast to the variants with semantic models (i.e. BCL variants). The lower performance at $t=2$ of \textit{attr-BImag} compared to \textit{class-BImag} highlights the limitations of attribute conditioning, probably due to a poorer VAE model when visual instances were already observed. Augmenting the condition with the class label (i.e. \textit{class-attr-BImag}) and the asymmetric approach \textit{asym-BImag} significantly alleviate this problem, both variants achieving the best performance in CUB 150/50 in AUTAC metric. In AwA \textit{class-attr-BImag} performs best in $t=1,2$ and also average AUTAC. In summary, BCL methods outperform CL (i.e. class-BImag) at initial times (thanks to the semantic model), while outperforming GZSL ($t=1$ row) by updating the visual model over time. Overall, properly using semantic information and class labels in our VAE component helps us to improve the functionality of both CL and GZSL.


\section{Conclusion}
We propose GCL as a novel and more realistic setting where continual learning is augmented with an explicit semantic model, which we argue is essential in humans to address visual recognition. We focus on the particular case of BCL, where the semantic model is fixed beforehand, but still generalizes (G)ZSL and CL.

We also propose the BImag framework based on feature generation in both forward and backward temporal directions, which we used to study the interplay between CL and semantic models. We observed that the semantic model may harms the ability to prevent forgetting. We propose two variants to alleviate this problem based on joint class-attribute conditioning and asymmetric generation.

\minisection{Acknowledgement}
We acknowledge the support from
Huawei Kirin Solution.

\bibliographystyle{icml2020}
\bibliography{bcl_v2}

\end{document}